\documentclass{article}

\usepackage[preprint]{neurips_2024}

\usepackage[utf8]{inputenc} 
\usepackage[T1]{fontenc}    
\usepackage{hyperref}       
\usepackage{url}            
\usepackage{booktabs}       
\usepackage{amsfonts}       
\usepackage{nicefrac}       
\usepackage{microtype}      
\usepackage{xcolor}         
\usepackage{qtree}
\usepackage{pgfplots}
\usepgfplotslibrary{groupplots}

\usepackage{amsmath}
\usepackage{amssymb}
\usepackage{mathtools}
\usepackage{amsthm}
\usepackage{comment}
\usepackage{enumitem}

\theoremstyle{plain}
\newtheorem{theorem}{Theorem}[section]

\theoremstyle{definition}
\newtheorem{definition}[theorem]{Definition}

\theoremstyle{remark}

\usepackage[textsize=tiny]{todonotes}

\newcommand{\add}[1]{#1}
\newif\ifremovemode
\removemodetrue 

\title{Global Lyapunov functions: a long-standing open problem in mathematics, with symbolic transformers}
\begin{document}

\author{Alberto Alfarano\thanks{Equal contribution} \\ FAIR, Meta\\ \texttt{albealfa@meta.com} \\
\And Fran\c{c}ois Charton\footnotemark[1]\\FAIR, Meta -- Ecole des Ponts \\ \texttt{fcharton@meta.com}
\And
Amaury Hayat\footnotemark[1] \\ CERMICS, Ecole des Ponts - Institut Polytechnique de Paris \\ \texttt{amaury.hayat@enpc.fr}}

\maketitle

\begin{abstract}
Despite their spectacular progress, language models still struggle on complex reasoning tasks, such as advanced mathematics.
We consider a long-standing open problem in mathematics: discovering a Lyapunov function that ensures the global stability of a dynamical system. This problem has no known general solution, and algorithmic solvers only exist for some small polynomial systems.
We propose a new method for generating synthetic training samples from random solutions, and show that sequence-to-sequence transformers trained on such datasets perform better than algorithmic solvers and humans on polynomial systems, and can discover new Lyapunov functions for non-polynomial systems.

\end{abstract}

\section{Introduction}

As large language models achieve human-level performance over a broad set of tasks~\citep{cicero2022,roziere2024code,zhou2023large}, their capability to \emph{reason} becomes a focus of discussion and research. There is no single definition of reasoning, and work in this area encompasses factuality, real world alignment, compositionality, the discovery and following of rules, \&c. Still, mathematics are considered as one of the purest, and most demanding, forms of reasoning~\citep{kant1787}. As such, solving research-level mathematical problems is a major milestone in demonstrating the reasoning capabilities of language models. Such an advance in AI would also transform mathematical practice.

There is little research on applying language models to open problems of mathematics. 
Except a few papers on combinatorial optimization and graph theory~\citep{funsearch2023,wagner2021constructions}, most prior works focus on problems with known solutions~\citep{trinh2024,HTPS,polu2020generative,charton2020learning}. 
We believe this lack of results is due to two main reasons. First, research problems may require specialized work by mathematicians \citep{Buzzard_2020} before they can be handed to language models. Second, most math transformers are trained on sets of problems and solutions which are hard to generate in the case of open problems, when no generic method for finding a solution is known. 

In this paper, we focus on a long-standing, yet easy to formalize, open problem in mathematics: 
discovering the Lyapunov functions that control the global stability of dynamical systems -- the boundedness of their solutions when time goes to infinity with respect to an equilibrium or an orbit. A famous instance of this problem is the \emph{three-body problem}: 
the long-term stability of a system of three celestial bodies subjected to gravitation. The stability problem was studied by Newton, Lagrange and Poincaré. Lyapunov discovered that stability is guaranteed if an entropy-like function for the system --the Lyapunov function-- can be found. Unfortunately, no
method is known for deriving Lyapunov functions in the general case, and Lyapunov functions are only known for a small number of systems.

We propose a new technique for generating training data from randomly sampled Lyapunov functions. Sequence-to-sequence transformers trained on these datasets achieve near perfect accuracy ($99\%$) on held-out test sets, and 
very high performance ($73\%$) on out-of-distribution test sets. We show that higher accuracies ($84\%$) can be achieved by enriching the training set with a small number ($300$) of easier examples that can be solved with existing algorithmic methods. These enriched models greatly outperform state-of-the-art techniques and human performance on a variety of benchmarks.

Finally, we test the capability of our models to discover yet unknown Lyapunov functions on randomly generated systems. On polynomial systems, the only ones current methods can solve, our models find Lyapunov function for $10.1\%$ or systems, vs $2.1\%$ for state-of-the-art techniques. On non-polynomial systems, where no algorithm is known, our best models discover new Lyapunov functions for $12.7\%$ of systems. 
Our research demonstrates that generative models can be used to solve research-level problems in mathematics, by providing mathematicians with guesses of possible solutions. 
The solutions proposed by the black-box model are explicit and their mathematical correctness can be verified. 
We believe this research is an AI-driven blueprint for solving open problems in mathematics.

\subsection*{Related works}
\label{app:relatedworks}

Most classical methods for finding Lyapunov rely on parameterized families of candidate solutions, and attempt to derive conditions on the parameters \citep{coron-book,giesl2007construction}. Additional techniques such as backstepping or forwarding \citep[Chap. 12]{coron-book} were introduced to leverage the specifics of particular systems. These techniques are limited to 
specific, or simple, systems. 
The global Lyapunov functions of polynomial systems that are sums of squares of polynomials of 
given degree can be found by computational-intensive algorithmic tools, such as  SOSTOOLS \citep{prajna2002introducing,prajna2005sostools}, which leverage the fact that the Lyapunov function belongs to a finite-dimensional space.       

Methods involving neural networks have been proposed in recent years \citep{NEURIPS2019_2647c1db, grande2023augmented, fossil2, lyznet}. They train 
feed-forward networks to approximate Lyapunov functions of a given system, and use a Satisfiability Modulo Theories (SMT) solver as a verifier which proposes potential counter-examples. This approach, very different from ours, was shown to be successful for
several well-studied high dimensional systems. However, it only finds local or semi-global Lyapunov functions (see Definition \ref{def:semiglobal}). Since the Lyapunov functions that are found are implicit, it would be hard for mathematicians to check whether they are global Lyapunov functions or not. Semi-global Lyapunov functions are useful in many engineering fields such as robotics, where one wants a system to be robust to small perturbations.
In other fields, like epidemics, being resilient to large perturbations
is central, and global Lyapunov functions are required.

Transformers trained on synthetic datasets have been proposed for many problems of mathematics, including arithmetic~\citep{arithmetic}, linear algebra~\citep{charton2022linear}, symbolic integration~\citep{lample2019}, symbolic regression~\citep{biggio2021neural}, Shortest Vector Problem \citep{wenger2022salsa}, Gröbner basis computation \cite{kera2023learning} and theorem proving~\citep{polu2020generative}. 
\cite{charton2020learning} investigate a problem related to ours: the local stability of dynamical systems. Different architectures were used to solve hard problems in combinatorial optimisation~\citep{funsearch2023}, and graph theory~\citep{wagner2021constructions}.

\section{System stability and Lyapunov functions}

The stability of dynamical systems is a hard mathematical question, which intrigued many generations of mathematicians, from Newton and Lagrange in the 18th century, to Poincaré in the 20th in the context of the three-body problem. The main mathematical tool for assessing stability was proposed by Lyapunov, who showed in 1892 that a system is stable if a decreasing entropy-like function --the Lyapunov function-- can be found \citep{Khalil,coron-book,Lyapunov}. Later, the existence of a Lyapunov function was shown to be a necessary condition for the stability of large classes of systems \citep{persidskii1937theorem,massera1949liapounoff,kellett2015classical}. Unfortunately, these very strong results provide no clue on how to find Lyapunov functions, or just proving their existence for a particular system. In fact, 130 years later, systematic derivations of global Lyapunov functions are only known in a few special cases, and their derivation in the general case remains a well-known open problem.

\begin{figure}[t]
\begin{minipage}{0.47\textwidth} 
\includegraphics[width=0.8\textwidth]{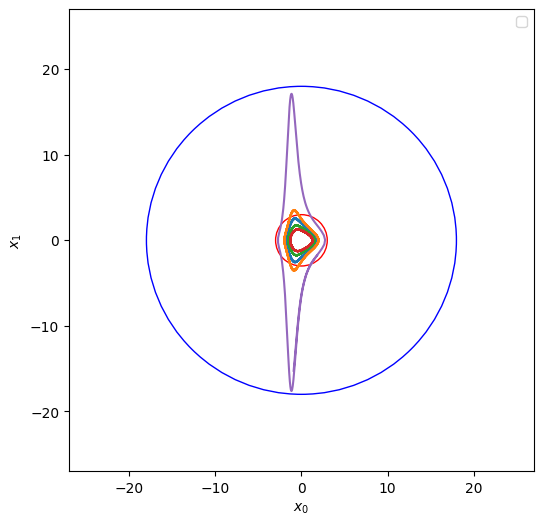}
\caption{\label{fig:stab} \small Dynamic of a stable system: trajectories may be complicated but as long as they start in the red ball they remain in the blue ball.}
\end{minipage}\hfill 
\begin{minipage}{0.50\textwidth}
\vspace{\baselineskip}
\begin{gather*}
\begin{cases}
    \dot x_{0} = -7x_0^5-4x_0^3x_1^2-5x_0^3\\
    \dot x_{1} = 7x_0^4-3x_1-2x_2\\
    \dot x_{2} = -8x_0^2-9x_2
\end{cases}\\
\begin{split}
V&(x) = 2x_{0}^{4} + 2x_{0}^{2}x_{1}^{2} + 3x_{0}^{2} + 2x_{1}^{2} + x_2^{2}
\end{split}
\end{gather*}
\begin{gather*}
\begin{cases}
\dot x_0 &= -x_0+x_0x_1 \\
\dot x_1 &= -x_1
\end{cases}\\
V(x) =\ln(1+5x_0^2) + x_1^2
\end{gather*}
\caption{\label{fig:lyap}\small Two stable systems and associated Lyapunov functions discovered by our model. The second, a polynomial system with a non-polynomial Lyapunov function, was studied  in~\cite{KrsticParrilo}.}
\end{minipage}
\end{figure}

In mathematical terms, we consider the dynamical system 
\begin{equation}
\label{eq:sys1}   
\dot x = f(x),
\end{equation}
where $x\in\mathbb{R}^{n}$, $f\in C^{1}(\mathbb{R}^{n})$ 
and $\dot x= \frac{dx}{dt}$. We want to know if the system has a stable equilibrium around a point $x^*$ such that $f(x^*)=0$. We assume, without loss of generality, that $x^*=0$.
\begin{definition}
    The system \eqref{eq:sys1} is \emph{stable} 
    when, for any $\varepsilon>0$, there exists $\eta>0$ such that, if $\|x(0)\|<\eta$, the system \eqref{eq:sys1} with initial condition $x(0)$ has a unique solution $x\in C^{1}([0,+\infty))$ and 
    \begin{equation}
    \|x(t)\|\leq \varepsilon,\;\;\forall\;t\in[0,+\infty).
    \end{equation}
\end{definition}
In other words, a system is stable if a solution that begins close to the origin ($\|x(0)\|<\eta$) stays close to the origin at all time ($\|x(t)\|\leq \varepsilon$). Lyapunov proved that the stability is related to the existence of what is now called a Lyapunov function.

\begin{definition}
The function $V\in C^{1}(\mathbb{R}^{n},\mathbb{R}_{+})$ is said to be a \emph{(global) Lyapunov function} for the system \eqref{eq:sys1} if the following condition are satisfied
\begin{equation}
\label{eq:defVcond}
\begin{split}
    V(0) = 0,\;\;\;\; &{
      \lim\limits_{\|x\|\rightarrow +\infty} V(x) = +\infty,}\\
    V(x)> 0,\;\;\;\; &\nabla V(x)\cdot f(x) \leq 0 \text{ for }x\neq 0.
\end{split}
\end{equation}
\end{definition}

\begin{theorem}[Lyapunov 1892]\label{lyapth}
If the system \eqref{eq:sys1} has a Lyapunov function, then it is stable.
\end{theorem}

In fact, the existence of a 
Lyapunov function is more powerful and provides additional information.
\begin{theorem}[LaSalle, 1961]
If the system \eqref{eq:sys1} has a Lyapunov function $V$, then all the solutions of \eqref{eq:sys1} converge to the largest invariant set of $\{f(x)\cdot \nabla V(x) = 0\}$.
\end{theorem}
In many cases this largest invariant set is reduced to $\{x^{*}=0\}$ and the system is said \emph{globally asymptotically stable} (all solutions converge to the equilibrium, see Appendix \ref{app:mathdef}).

Most dynamical systems are unstable. For instance, the solutions of the simple system $\dot x(t)= x(t)$ grow exponentially with time, and the solutions of $\dot x(t) = 1+x(t)^{2}$ ($x\in\mathbb{R}$) always blow up before $t=\pi$. No Lyapunov functions can be found for these systems.

On the other hand, stable systems can have an infinite number of Lyapunov functions. The system 
\begin{equation*}
\begin{cases}
    \dot x_0(t)= -x_0(t)\\
    \dot x_1(t)= -x_1(t)
\end{cases}
\end{equation*}
has $V(x) = a_0 x_0^2 + a_1 x_1^2$ as a Lyapunov function for any choice of $a_0 > 0$ and $a_1 > 0$. 

In the general case, there is no systematic way of discovering a Lyapunov function, or even showing that one exist. Tools exist for small polynomial systems with special ``sum of squares'' (SOS) Lyapunov functions, but they need a lot of resources, do not always find a solution, and fail once the systems involve more than a few variables.

We also consider a related, but easier, problem: finding nontrivial $V$ which are semi-definite positive, i.e. $V$ verifying $V(x) \geq 0$ instead of $V(x)>0$ in Equation~\eqref{eq:defVcond}. These functions, called \emph{barrier functions},
form ``barriers'' that 
divide $\mathbb{R}^{n}$ into two subspaces. A solution starting inside the barrier 
must remains in the same subspace, which is an invariant set of the system~\citep{prajna2007framework,xu2015robustness}. For polynomial systems, barrier functions are slightly easier to find using SOS solvers.

\section{Experimental settings}

In this work, we train sequence-to-sequence transformers~\citep{vaswani2017attention} to predict a Lyapunov function for a given system, when it exists. We frame the problem as a translation task: problems and solutions are represented as sequences of symbolic tokens, and the model is trained from generated pairs of systems and Lyapunov functions to minimize the cross-entropy between the predicted sequence and the correct solution. 
We train transformers with $8$ layers, $10$ attention heads and an embedding dimension of $640$ (ablation studies on different model sizes can be found in Appendix~\ref{app:suppl_results}), on batches of $16$ examples, using the Adam optimizer~\citep{kingma2014adam} with a learning rate of $10^{-4}$, an initial linear warm-up phase of 10,000 optimization steps, and inverse square root scheduling. All experiments run on $8$ V100 GPU with $32$ GB of memory, for $3$ or $4$ epochs of $2.4$ million examples per epoch. Training time is between 12 to 15 hours per GPU.

\textbf{Tokenization.} Model inputs are systems of the form $(\dot x_i = f_i(x_1, \dots ,x_n))_{i\in \{1,\dots ,n\}}$, represented by the $n$ functions $f_i$. Model outputs are single functions $V(x_1, \dots ,x_n)$. As in~\cite{lample2019}, functions are represented as trees, with operators in their internal nodes, and variables or constants as their leaves. Trees are then enumerated in Polish (pre-order) notation to produce sequences of tokens that can be processed by the transformer. 

All operators and variables are tokenized as single symbols (e.g. ‘\texttt{cos}' or ‘\texttt{$x_1$}'). Integer constants are tokenized as sequences of ``digits'' in base $1000$ (e.g. $1024$ as the sequence $[\texttt{+, 1, 24}]$), and real constants, in scientific notation, as pairs of two integers (mantissa and exponent, e.g. $-3.14$ as $[\texttt{-},314,\texttt{10\^{}}, \texttt{-}, 2]$). For instance:

\begin{minipage}{0.3\textwidth}
\begin{equation*}
\left\{\begin{split}
    \dot x_0 &= \cos(2.1 x_0) (x_1+2)\\
    \dot x_1 &= \sin(3x_1+2)
\end{split}\right. \quad\quad\text{ is represented as }
\end{equation*}
\end{minipage}\hfill
\begin{minipage}{0.5\textwidth}
\Tree [.*  [.cos [.* $2.1$ $x_0$ ]  ] [.+ $x_1$ $2$ ] ] \Tree [.sin [.+ [.* $3$ $x_1$ ] 2 ] ]
\end{minipage}

enumerated as the sequences: $[*, \cos, *, 2.1, x_0, +, x_1, 2]$ and $[\sin, +, *, 3, x_1, 2]$, and finally tokenized as 
[\texttt{*}, \texttt{cos}, \texttt{*}, \texttt{21}, \texttt{10\^{}}, \texttt{-}, \texttt{1}, $x_0$, \texttt{+}, $x_1$, \texttt{2}, \texttt{SEP}, \texttt{sin}, \texttt{+}, \texttt{*}, \texttt{3}, $x_1$, \texttt{2}] (using \texttt{SEP} as a separator).

\textbf{Evaluation.} Trained models are evaluated on sets of stable systems. Since systems have an infinite number of Lyapunov functions, we cannot check the model predictions by comparing them to the solutions from the test set, and need to use an external verifier. 
For polynomial systems, we verify that there exists a small positive polynomial $P$ such that $- \nabla V\cdot f$ and $V-P$ are sum of squares (SOS) of polynomials (with $P=0$ for barrier functions), using a Python solver based on SumOfSquares~\citep{Yuan2024}. 
For non-polynomial systems, we also use a verifier based on \texttt{shgo} that checks~\eqref{eq:defVcond} numerically. To further ensure correctness we also verify the symbolic solutions using Satisfiability Modulo Theories (SMT) solvers, relying on dReal ~\citep{dReal} for verification through interval analysis. This guarantees that equations~\eqref{eq:defVcond} hold, at least in a chosen ball around the origin. The performances of the two verifiers (numerical solver and SMT) are similar, a comparison is provided in Table~\ref{tab:smttimeout}. 
Both the SOS and SMT verifiers sometimes fail to return an answer. In that case, we classify the solution as wrong, even though it might have been correct. As a result, model accuracies may be underestimated.

Model predictions use beam search with early stopping, normalizing log-likelihood scores by their sequence length. We report results with beam size 1 (greedy decoding) and beam size $50$. With beam size $50$, we consider the model to be correct if one Lyapunov function is found among the $50$ guesses.

\section{Data generation}
\label{chap:sett}

Our models are trained and tested on large datasets of pairs of stable systems and associated Lyapunov functions. Sampling such stable systems raises two difficulties. First, most dynamical systems are unstable, and no general method exists for deciding whether a system is stable. Second, once a stable system is sampled, there is no general technique for finding a Lyapunov function, except in particular cases. In this paper, we rely on \textbf{Backward generation}~\citep{lample2019}, sampling solutions and generating associated problems, for the general case, and \textbf{forward generation}, sampling systems and calculating their solutions with a solver, for the tractable polynomial systems of small degree.

\subsection{Backward generation}\label{sec:backgen}

Backward generation methods, sampling problems from their solutions, are only useful if the model can be prevented from learning to reverse the generation procedure, or from ``reading'' the solutions in the generated problems. For instance, when training a model to solve the hard problem of finding the roots of an integer polynomial~\citep{rootscharton}, one can easily generate a polynomial from its roots, i.e. from the roots $3, 5$ and $7$, generate the polynomial: \[P(X) = 2(X^2+1)(X-3)(X-5)(X-7).\] 
However, if the model is trained from factorized form of $P(X)$, it will learn to read the roots in the problem, instead of computing them. On the other hand, the developed and simplified form \[P(X) = 2X^5-30X^4+144X^3-240X^2+142X-210\] offers no clues. 
A second difficulty of backward generation is that sampling solutions instead of problems biases the training distribution. A model trained on backward-generated data may not perform well on a forward-generated test set. Finally, prior work~\cite{yehuda2020} observed that, for hard problems, backward generation methods sometimes focus on easier sub-problems (see, for instance, our comment below about choosing  $f = - \nabla V$ in step $2$). 

We propose a procedure for generating a stable system $S$ from a random Lyapunov function $V$. The rationale is the following. Since $V$ must be positive with a strict minimum in $0$, and tend to infinity at infinity (\eqref{eq:defVcond}), we first generate $V=V_{\textrm{proper}} + V_{\textrm{cross}}$ where $V_{\textrm{proper}}$ belongs to a class of functions with a guaranteed strict minimum in zero and $V_{\textrm{cross}}$ to a larger class of non-negative functions, valued 0 at the origin, but with no guarantee of a strict minimum (step 1 and Appendix \ref{app:generation}).
 From $V$, we need to generate $f$ so that the third condition of~\eqref{eq:defVcond} is met. A naive solution would be $f = -\nabla V$ since $f\cdot \nabla V \leq 0$ would hold. But this would severely limit the systems we create, and turn the Lyapunov function discovery problem (find $V$ from $f$) into an easier integration problem (find $V$ from $-\nabla V$). Instead, starting from $f_{0} = -\nabla V$, we apply the following transformations:

\begin{itemize}[nosep]
    \item multiply each coordinate of $f_{0}$ by random non-negative functions $h_i^2$ (step 4) and call it $\tilde{f}_{0}$. 
    \item generate a random function $\phi = \sum_{i=1}^p g_{i}(x)e^{i}(x)$ (steps 2 and 3),  where $e^i$ are orthogonal to $\nabla V(x)$, and set $f=\varphi+\tilde{f}_{0}$. We have $\phi\cdot \nabla V = 0$ and $(\phi+\tilde{f}_{0})\cdot \nabla V\leq 0$.
\end{itemize}

These transformations guarantee that all conditions in~\eqref{eq:defVcond} are met. On the other hand, they allow $f$ to span a very large set of systems, since any $f$ satisfying $\nabla V(x)\cdot f(x)\leq 0$ can be written as the sum of a function collinear to $\nabla V(x)$ and a function orthogonal to $\nabla V(x)$. 

Specifically, the procedure can be summarized as follows (see Appendix~\ref{app:generation} for more details).

\begin{itemize}[nosep]
    \item[] \textbf{Step 1} Generate a random function $V$, satisfying $V(x)>V(0),\;\;\forall x\in \mathbb{R}^{n}\setminus\{0\}$, and $
    V(x)\rightarrow +\infty
    $ when $\|x\|\rightarrow +\infty$. 
    \item[]\textbf{Step 2} Compute the gradient $\nabla V(x)$ and denote $\mathcal{H}_{x}=\{z\in\mathbb{R}^{n}\;|\; z\cdot \nabla V(x)=0\}$ the hyperplane\footnote{if $\nabla V(x) = 0$ this is the whole space instead, but this does not change the method.} orthogonal to $\nabla V(x)$, for any $x\in\mathbb{R}^{n}$.
    \item[]\textbf{Step 3} Select $1\leq p \leq n$ at random and sample $p$ vectors $\{e^{i}(x)\}_{i\in\{1,...,p\}}$ from hyperplane $\mathcal{H}_{x}$. Generate $p$ real-valued functions $(g_{i})_{i\in\{1,...,p\}}$.
\item[]\textbf{Step 4} Select $1<k_{1}\leq n$ at random, generate $k_1$ random real-valued functions $(h_{i})_{i\in\{1,...,k_{1}\}}$, set $h_{i}= 0$ for $ k_{1}+1\leq i\leq n$. 
\item[]\textbf{Step 5} Build the $n$ functions
\[f(x) = -(h_{\pi(i)}^{2}(x)(\nabla V)_i(x))_{i\in\{1,...,n\}} + \sum\limits_{i=1}^{p}g_{i}(x)e^{i}(x),\]
with $\pi$ a random permutation of $\{1,...,n\}$. 
\item[]\textbf{Step 6} Simplify the functions $f_i$, obscuring patterns from the generative process.
\end{itemize}

This method produces a stable system $S: \dot x = f(x)$, with $V$ as its Lyapunov function. The difficulty of inferring $V$ from $S$ hinges on a careful choice of the vectors $e^{i}$. For instance, if we naively select $e^i$ as an orthonormal basis of $\mathcal{H}_{x}$, computed from $\nabla V(x)$ by Gram-Schmidt orthogonalization, prefactors like $1/\|\nabla V(x)\|$ appear at step 3, and are unlikely to simplify away at step 6. This provides the model with a shortcut: reading $\|\nabla V(x)\|$ in $S$, and using it to recover $\nabla V$ and then $V$, not a trivial task, but an easier one than discovering Lyapunov functions. To counter this, we relax the orthonormality condition on $e^i(x)$, so that $1/\|\nabla V(x) \|$ never appears, yet keep the $e^i(x)$ simple enough for $\nabla V$-specific patterns in $\sum_i {g_i(x) e^i (x)}$ to simplify away at step 6. 
We also want to ensure that the $e^i$ span all of $\mathcal{H}_{x}$, or the systems generated will not be diverse enough.

In our experiments, we slightly modify this procedure, by running steps $2$ to $6$ five times for each Lyapunov function $V$ created at step $1$. As a result, $5$ systems are generated that share the same Lyapunov function (a discussion of this choice can be found in Appendix~\ref{sec:multigen}). From a mathematical point of view, a Lyapunov function describes a hidden quantity in a system, and we believe that providing the model with several systems that share this hidden quantity should help it learn the parts of the system that contribute to this hidden quantity, and therefore learn a Lyapunov function.

This procedure can be tuned to generate specific classes of systems. By choosing $V$, $g_i$ and $h_i$ in particular classes, we can constrain the system functions $f_i$ to be polynomials, polynomials of functions (e.g. trigonometric polynomials), or more general functions (see Appendix~\ref{app:design_params} for more).

The Lyapunov functions obtained here are correct by design. Nevertheless, we still performed an evaluation of the solutions both as a safeguard and to benchmark the failure and timeout rates of the SMT and SOS solvers on correct solutions, which we report in Table ~\ref{tab:smttimeout}.

\begin{table}[t]
\small
\centering
\begin{tabular}{l|cc|cc}
    \toprule
     & \multicolumn{2}{|c}{SMT Solver} & \multicolumn{2}{|c}{SOS Solver} \\
    Timeout &10 minutes &60 minutes&10 minutes&60 minutes\\
    \midrule
    Correct Lyap function & 82.6 & 94.1 & 89.6 & 95.3\\
    Solver Timeouts & 17.4 & 5.9 & 10.4 & 4.7\\
    Incorrect Lyap function & 0 & 0 & 0 & 0 \\
    \bottomrule
\end{tabular}
\caption{\small \textbf{SMT and SOS timeout and error rates,} benchmarked on correct Lyapunov functions.}
\label{tab:smttimeout}
\end{table}

\subsection{Forward generation}\label{sec:genpoly}
Whereas the stability problem is unsolved in the general case, methods exist to calculate Lyapunov functions of polynomial systems, when they exist and can be written as a sum of squares of polynomials (see Section~\ref{app:relatedworks}). These algorithms, of polynomial complexity, are very efficient for small systems, but their CPU and memory requirements explode as the size of the systems grows. We leverage them to generate forward datasets, as follows.
\begin{enumerate}[nosep]
\item[]\textbf{Step 1} Generate a polynomial system at random
\item[]\textbf{Step 2} Use a routine to find a polynomial sum-of-squares (SOS) Lyapunov function.
\item[]\textbf{Step 3} Keep the system if such function exists, restart from step 1 otherwise.
\end{enumerate}
This approach has several limitations. First, since most polynomial systems are not stable, and the computation of SOS Lyapunov function involves a complicated search~\citep{prajna2005sostools}, it is slow and limited to small systems of polynomials with small degree. Second, because not all stable polynomial systems have polynomial SOS Lyapunov functions~\citep{KrsticParrilo}, it can only generate a subset of stable polynomial systems.

Finally, SOS routines process the constraints in Equation~\eqref{eq:defVcond} by solving semi-definite programming (SDP) problems. This guarantees that $V$ is a sum-of-squares, hence we have $V(x)\geq 0$, but not necessarily $V(x)>0$, for $x\neq 0$. As a result, these methods can only discover \emph{barrier functions}. State-of-the-art methods circumvent this by introducing the stronger constraint $V(x)\geq \sum_{i=1}^{n}\varepsilon_{i}x_{i}^{2}$, 
with $\varepsilon_{i}$ small \cite{prajna2002introducing}. $V$ then has a unique minimum in $x=0$, which makes it a Lyapunov function, but this further restricts the class of polynomial systems that the method can solve.

\subsection{Datasets} 

We generate $2$ backward and $2$ forward datasets for training and evaluation purpose, and one smaller forward dataset for evaluation purposes (see Table~\ref{tab:datasets} in Appendix~\ref{app:datasets} for a list). 

\textbf{Backward datasets} Our main backward set, \textbf{BPoly}, features $1$ million \add{non-degenerate} polynomial systems $S$ with integer coefficients, and $2$ to $5$ equations (in equal proportions). 
We also create \textbf{BNonPoly}, a dataset of $1$ million non-degenerate non-polynomial systems with $2$ to $5$ equations. In this dataset, the coordinates of $f$ are polynomials of general functions, e.g. trigonometric polynomials, or functions such as $3\cos(x_1)+ 2 x_1 e^{x_2}$. For such general systems, no method for discovering a Lyapunov function is known.

\textbf{Forward datasets} All $2$ forward datasets are generated using a solver derived from the SumOfSquares package in Python, and implementing techniques similar to those used in SOSTOOLS (see Appendix~\ref{sec:SOS}). All systems in these datasets are non-zero integer polynomials with $2$ to $3$ equations, and integer polynomial Lyapunov functions -- the only systems these methods can solve. We create \textbf{FLyap}, a dataset of 100,000 systems having a non-homogeneous polynomial as a Lyapunov function. We also have a dataset focusing on barrier functions (see the end of section~\ref{sec:genpoly}): \textbf{FBarr} features 300,000 systems having a non-homogeneous polynomial as a barrier function.
The small size of these datasets is due to the computational cost of SOS methods, and the difficulty of discovering Lyapunov or barrier functions. 

To allow for comparison with SOSTOOL, the state-of-the-art method for discovering Lyapunov functions of polynomial systems, we also generated a test set of 1,500 polynomial systems with integer coefficients that SOSTOOLS can solve (\textbf{FSOSTOOLS}).

\section{Results}
\label{main_results}
Our models trained on different datasets achieve near perfect accuracy on held-out test sets, and very high performances on out-of-distribution test sets,
especially when enriching the training set with a small number of forward examples.
They greatly outperform state-of-the-art techniques 
and also allow to discover Lyapunov functions for new systems. 
These results are detailed below. 

\subsection{In and out-of-distribution accuracy}
\label{sec:indomain}
In this section, we present the performance of models trained on the $4$ datasets. All models achieve high in-domain accuracy -- when tested on held-out test sets from the datasets they were trained on (Table~\ref{tab:indomain}). On the forward datasets, barrier functions are predicted with more than $90\%$ accuracy, and Lyapunov functions with more than $80\%$. On backward datasets, models trained on BPoly achieve close to $100\%$ accuracy. We note that beam search, i.e. allowing several guesses at the solution, brings a significant increase in performance ($7$ to $10\%$ with beam size $50$, for the low-performing models). We use beam size $50$ in all further experiments.

\begin{table}[h]

\centering
\small
\begin{tabular}{lccc||lcc}
\toprule
 &\multicolumn{2}{c}{Accuracy}&& &\multicolumn{2}{c}{Accuracy}\\
Backward datasets& bs=1 & bs=50 &&Forward datasets & bs=1 & bs=50\\
\midrule
BPoly (polynomial) & 99 & 100 && FBarr (barrier) & 93 &98 \\
BNonPoly (non-poly) & 77 & 87 && FLyap (Lyapunov) & 81 &88 \\
\bottomrule
\end{tabular}
\caption{\small \textbf{In-domain accuracy of models}. Beam size (bs) 1 and 50.}
\label{tab:indomain}
\end{table}

The litmus test for models trained on generated data is their ability to generalize out-of-distribution (OOD). 
Table~\ref{tab:outdomain} presents evaluations of backward models on forward-generated sets (and the other way around). All backward models achieve high accuracy ($73$ to $75\%$) when tested on forward-generated random polynomial systems with a sum-of-squares Lyapunov functions (FLyap). The best performances are achieved by non-polynomial systems (BNonPoly), the most diverse training set. 
The lower accuracy of backward models on forward-generated sets of systems with barrier functions (FBarr) may be due to the fact that many barrier functions are not necessarily Lyapunov functions. On those test sets, backward models must cope with a different distribution and a (slightly) different task. Forward models, on the other hand, achieve low performance on backward test sets. This is possibly due to the small size of these training set.

Overall, these results seem to confirm that backward-trained models are not learning to invert their generative procedure. If it were the case, their performance on the forward test sets would be close to zero. They also display good OOD accuracy.

\begin{table}[h]

\centering
\small
\begin{tabular}{lcc||lc}
\toprule
Backward datasets & FLyap & FBarr  & Forward datasets & BPoly \\
\midrule
BPoly (polynomial) & \textbf{73} & 35  & FBarr (barrier) & 15 \\
BNonPoly (non-poly) & \textbf{75} & 24  & FLyap (Lyapunov) & 10 \\
\bottomrule
\end{tabular}
\caption{\small \textbf{Out-of-domain accuracy of models}. Beam size 50. Columns are the test sets.}
\label{tab:outdomain}
\end{table}

\subsection{Enriching training distributions for improved performance}
\label{sec:enriching}
To improve the OOD performance of backward models, we add to their training set a tiny number of forward-generated examples, as in~\cite{jelassi2023length}. Interestingly, this brings a significant increase in performance (Table~\ref{tab:mixing}). Adding $300$ examples from FBarr to BPoly brings accuracy on FBarr from $35$ to $89\%$ (even though the proportion  of forward examples in the training set is only $0.03\%$) and increases OOD accuracy on FLyap by more than $10$ points. Adding examples from FLyap brings less improvement. 

These results indicate that the OOD performance of models trained on backward-generated data can be greatly improved by adding to the training set a small number of examples (tens or hundreds) that we know how to solve. Here, the additional examples solve a weaker but related problem: discovering barrier functions. The small number of examples needed to boost performance makes this technique especially cost-effective. 


\begin{table}[h]
\centering
\small
\begin{tabular}{lc|cc}
\toprule
Forward & Examples added & & \\
datasets & (1M in training set) & FLyap & FBarr \\
\midrule
No mixture & 0 & 73 & 35 \\
\midrule
FBarr & 30 & 75 & 61 \\
 & 300 & 83 & 89 \\
 & 3,000 & 85 & 93 \\
 & 30,000 & 89 & 95 \\
\midrule
FLyap & 10 & 75 & 25 \\
& 100 & 82 & 29 \\
& 1,000 & 83 & 37 \\
& 10,000 & 86 & 38 \\
\bottomrule
\end{tabular}
\caption{\small \textbf{Mixing backward data (BPoly) with a small number of forward examples}. Beam size 50. }
\label{tab:mixing}
\end{table}


\subsection{Comparing with state-of-the-art baselines}
\label{sec:comparing}
To provide a baseline for our models, we developed \texttt{findlyap}, a Python counterpart to the MATLAB Lyapunov function finder from SOSTOOLS (see Appendix \ref{sec:SOS}). We also introduce FSOSTOOLS, a test set of 1,500 polynomial systems with integer coefficients that SOSTOOLS can solve. 
We also tested AI-based tools (see Appendix ~\ref{app:sweep} for the full list of parameters sweeps we used for each of these methods),
such as Fossil 2~\citep{fossil2}, ANLC v2~\citep{grande2023augmented} and LyzNet~\citep{lyznet}. These methods achieve low accuracies on our test sets. 
This might be due to the fact that these tools are designed to solve a different problem: discovering local or semi-global Lyapunov function (and potentially finding a control function), while we target global Lyapunov functions.


\begin{table}[h]
\centering
\small
\begin{tabular}{l|c|ccc|cccc}
\toprule
& SOSTOOL & \multicolumn{3}{|c}{Existing AI methods} & \multicolumn{4}{|c}{Models}\\
Test sets & \texttt{findlyap} & Fossil 2 & ANLC & LyzNet & PolyMixture & FBarr & FLyap & BPoly  \\
\midrule
FSOSTOOLS & - & 32 & 30 & 46 & \textbf{84} & 80 & 53 & 54 \\
FBarr  & -& 12 & 18 & 28 & \textbf{89} & - & 28 & 35\\
FLyap & - & 42 & 32 & 66 & 83 & \textbf{93} & - & 73\\
BPoly & 15 & 10 & 6 & 24 & \textbf{99}  & 15 & 10 & - \\
\bottomrule
\end{tabular}
\caption{\small \textbf{Performance comparison on different test sets}. Beam size 50. PolyMixture is BPoly + 300 FBarr. }
\label{tab:comparison_state_art}
\end{table}

Table~\ref{tab:comparison_state_art} compares \texttt{findlyap} and AI-based tools to our models on all available test sets.  A model trained on BPoly complemented with $500$ systems from FBarr (PolyMixture) achieves $84\%$ on FSOSTOOLS, confirming the high OOD accuracy of mixture models. On all generated test sets, PolyMixture achieves accuracies over $84\%$ whereas \texttt{findlyap} achieves $15\%$ on the backward generated test set.
This demonstrates that, on polynomial systems, transformers trained from backward-generated data achieve very strong results compared to the previous state of the art.

On average Transformer-based models are also much faster than SOS methods. When trying to solve a random polynomial system with 2 to 5 equations (as used in Section \ref{sec:intothewild}), 
\texttt{findlyap} takes an average of 935.2s (with a timeout of 2400s).
For our models, inference and verification of one system takes 2.6s on average with greedy decoding, and 13.9s with beam size $50$.

\subsection{Into the wild - discovering new mathematics}
\label{sec:intothewild}
Our ultimate goal is to discover new Lyapunov functions.
To test our models' ability to do so, we generate three datasets of random systems: polynomials systems with $2$ or $3$ equations (\textbf{Poly3}), polynomial systems with $2$ to $5$ equations (\textbf{Poly5}), and non-polynomial systems with $2$ or $3$ equations (\textbf{NonPoly}). For each dataset, we generate 100,000 random systems and eliminate those that are trivially locally exponentially unstable in $x^*=0$, because the Jacobian of the system has an
eigenvalue with strictly positive real part~\citep{Khalil}. We compare \texttt{findlyap} and AI based methods
with two models trained on polynomial systems, FBarr, and PolyM(ixture) -- a mixture of BPoly and 300 examples from FBarr-- and one model trained on a mixture of BPoly, BNonPoly and 300 examples from FBarr (NonPolyM).

Table~\ref{tab:intothewild} presents the percentage of correct solutions found by our models. On the polynomial datasets, our best model (PolyM) discover Lyapunov functions for $11.8$ and $10.1\%$ of the (degree 3 and degree 5) systems, ten times more than \texttt{findlyap}. For non-polynomial systems, Lyapunov functions are found for $12.7\%$ of examples. These results demonstrate that language model trained from generated datasets of systems and Lyapunov function can indeed discover yet unknown Lyapunov functions 
and perform at a much higher level that state-of-the-art SOS solvers. 

\begin{table}[h]
\centering
\small
\begin{tabular}{lc|c|ccc|c|cc}
\toprule
& Sample & SOSTOOL & \multicolumn{3}{c|}{Existing AI methods} & Forward & \multicolumn{2}{c}{Backward models}\\
Test set & size & \texttt{findlyap} & Fossil 2 & ANLC & LyzNet & FBarr & PolyM & NonPolyM \\
\midrule
Poly3 & 65,215 & 1.1 & 0.9 & 0.6 & 4.3 & 11.7 & \textbf{11.8} & 11.2 \\
Poly5 & 60,412 & 0.7 & 0.3 & 0.2 & 2.1 & 8.0 & \textbf{10.1} & 9.9 \\
NonPoly & 19,746 & - & 1.0 & 0.6 & 3.5 & - & - & \textbf{12.7} \\
\bottomrule
\end{tabular}
\caption{\small \textbf{Discovering Lyapunov comparison for random systems}. Beam size 50. PolyM is BPoly + 300 FBarr. NonPolyM is BNonPoly + BPoly + 300 FBarr. 
}
\label{tab:intothewild}
\end{table}


\subsection{Expert iteration}
\label{sec:expertiteration}

\add{Given the performance on our model in Table ~\ref{tab:intothewild}, we can use the newly solved problems to further fine-tune the model. Specifically, we create a sample of verified model predictions for polynomial systems, \textbf{FIntoTheWild}, we add it to the original training sample and we continue training the model.} 

\add{We test different strategy to finetune the model and we report performance on forward benchmarks and ``into the wild'' in Table ~\ref{tab:expert}.}
\begin{enumerate}[nosep]
    \item[\textit{n1}:] \add{Add 20,600 samples from BPoly (20,000), FBarr (50), FLyap (50) and FIntoTheWild (500) to keep similar proportion used during pretraining}
    \item[\textit{n2}:] \add{Add 2,000 samples from FLyap (1,000) and FIntoTheWild (1,000) to improve on both forward benchmark and in the wild}
    \item[\textit{n3}:] \add{Add 50 samples from FIntoTheWild to show that this indeed helps}
    \item[\textit{n4}:] \add{Add 1,000 samples from FIntoTheWild}
    \item[\textit{n5}:] \add{Add 2,000 samples from FIntoTheWild}
    \item[\textit{n6}:] \add{Add 5,000 samples from FIntoTheWild to see if there are benefits to add more samples}
\end{enumerate}

\add{We also retrain a model (\textit{n7}) from scratch using a mixture of BPoly (1M), FBarr (500), FLyap (500) and FIntoTheWild (2,000).}

\begin{table}[h]
\centering
\small
\begin{tabular}{c|cc|cc}
\toprule
& \multicolumn{2}{|c}{Forward benchmark} & \multicolumn{2}{|c}{Regenerated IntoTheWild} \\
Strategy & FBarr & FLyap & Poly3 & Poly5 \\
\midrule
Baseline & 93 & 84 & 11.7 & 9.6 \\
\midrule
\textit{n1} & 94 & 84 & 10.3 & 9.6 \\
\textit{n2} & 90 & 85 & 12.2 & 11.3 \\
\textit{n3} & 92 & 84 & 12.4 & 10.1 \\
\textit{n4} & 92 & 84 & \textbf{13.5} & \textbf{11.9} \\
\textit{n5} & 89 & 79 & 13.5 & 11.9 \\
\textit{n6} & 85 & 72 & 13.5 & 11.9 \\
\textit{n7} & 93 & 81 & 12.1 & 10.0 \\
\bottomrule
\end{tabular}
\caption{\small \textbf{Expert iteration using IntoTheWild correct guesses}. The Poly3 and Poly5 test sets are regenerated, to prevent data contamination. 
}
\label{tab:expert}
\end{table}
We notice that the addition of 1,000 verified predictions to our training set of 1 million improves performance on the ``into to wild'' test sets by about 15\%, while not affecting the other test sets (\textit{n4}). Adding more examples seems to be detrimental, as it decreases the performance on other benchmarks (\textit{n5} and \textit{n6}). We also notice that finetuning with mixed data from other distributions is not efficient (\textit{n1} and \textit{n2}) and a small contribution already help to get some improvements (result \textit{n3}). Finally, it's not efficient to pretrain the model from scratch using data from FIntoTheWild (\textit{n7}).

\section{Discussion}

We have shown that models can be trained from generated datasets to solve a long-standing open problem in mathematics: discovering the Lyapunov functions of stable dynamical systems. 
For random polynomial systems, our best models can discover Lyapunov functions in five times more cases than state-of-the-art methods. They can also discover Lyapunov functions of non-polynomial systems, for which no algorithm is yet known, and were able to re-discover a non-polynomial Lyapunov function of a polynomial systems discovered by~\cite{KrsticParrilo} (Appendix \ref{app:examples}).

The backward generation method introduced in section~\ref{sec:backgen} is the key innovation in this paper. The main problem with such approaches is their tendency to generate training sets with very specific distributions, which prevent models from generalizing to general instances of the problem. Our models can generalize out of their training distributions (Table~\ref{tab:outdomain}), and we can improve their performance by adding to their training set a tiny number of systems that we know how to solve (Table~\ref{tab:comparison_state_art}).

While our models exceed the algorithmic state of the art, one might wonder \textbf{how they compare to human mathematicians}. To this effect, we proposed $75$ problems from the FSOSTOOLS dataset (polynomial systems with $2$ or $3$ equations) as an examination for 25 first year Masters students in mathematics, following a course on the subject. Each student was given 3 systems chosen at random 
and had a total of 30 min. Their performance was 9.33\%, significantly lower than our models ($84\%$).  

Our work has a number of limitations. Because there is no known way to tell whether a random system is stable, we lack a good benchmark on non-polynomial systems. Also, all the systems studied in this paper are relatively small, at most $5$ equations for polynomial systems and $3$ for non-polynomial. We believe that scaling to larger models should help tackle larger, and more complex, systems. Finally, this work could be extended to take into account the domain of definition of non-polynomial systems.

The broader implications of our work extend into two directions: the capability of transformers to reason, and the potential role of AI in scientific discovery.
While large language models perform at human level on a broad set of tasks, they are still embarrassingly clumsy on many simple problems of logic and reasoning, to the point that it was suggested that planning and high level reasoning may be an inherent limitation of auto-regressive transformer architectures.
\add{Our results suggest that transformers can indeed be trained to discover solutions to a hard problem of symbolic mathematics that humans solve through reasoning, and that this is enabled by a careful selection of training examples, instead of a change of architecture. We do not claim that the Transformer is reasoning but it may instead solve the problem by a kind of ``super-intuition'' that stems from a deep understanding of a mathematical problem.}

From a mathematical point of view, we propose a new, AI-based, procedure for finding Lyapunov functions, for a broader class of systems than were previously solvable using current mathematical theories. 
While this systematic procedure remains a black box, and the ``thought process'' of the transformer cannot be elucidated, the solutions are explicit and their mathematical correctness can be verified. This suggests that generative models can already be used to solve research-level problems in mathematics, by providing mathematicians with guesses of possible solutions. While a small minority of  
mathematicians is currently using deep-learning tools, we believe generative models have the potential to foster tremendous progress on a number of research subjects, and may eventually become a central component in the future landscape of mathematical practice.

\subsection*{Acknowledgements}
This work was performed in part using HPC resources from GENCI–IDRIS (Grant 2023-AD011014527). The authors also acknowledge the Office of Advanced Research Computing (OARC) at Rutgers, The State University of New Jersey. 
The authors would also like to thank the Master students of the Mathematics and Computer Science Department of the Ecole des Ponts - IP Paris from the year 2023-2024 who attended the course Control of Dynamical Systems.

\bibliography{lyapunov}
\bibliographystyle{plainnat}

\newpage
\appendix
\section*{Appendix}

{
\section{Mathematical definitions}
\label{app:mathdef}

In this Appendix, we recall several mathematical definitions and theorems related to the Lyapunov function problem. We first introduce the notion of global asymptotic stability (GAS).

\begin{definition}
We say that the (equilibrium $x^{*}=0$ of the) system \eqref{eq:sys1} is \emph{globally asymptotically stable} if it is stable and for any $x_{0}\in \mathbb{R}^{n}$ there exists a unique solution $x\in C^{1}([0,+\infty);\mathbb{R}^{n})$ to \eqref{eq:sys1} which satisfies in addition
\begin{equation}
\lim\limits_{t\rightarrow+\infty} x(t) = 0.
\end{equation}
\end{definition}
This notion translates the fact that the equilibrium $x^{*}=0$ is robust even to large perturbations. This notion is related to the existence of a Lyapunov function thanks, for instance, to LaSalle Invariance Principle:
\begin{theorem}[LaSalle Invariance Principle (global)]
Assume there exists a Lyapunov function for the system \eqref{eq:sys1} and let $S$ be the largest subset of $\{\nabla V(x)\cdot f(x) = 0\}$ that is invariant by the dynamics of \eqref{eq:sys1}. If $S = \{0\}$, then the system \eqref{eq:sys1} is globally asymptotically stable.
\end{theorem}
Note that if $\nabla V(x)\cdot f(x) <0$ for any $x\neq 0$ then necessarily $S=\{0\}$. Because finding a (global) Lyapunov function is a challenging mathematical problem, and still an open problem in general, weaker notions exists.
\begin{definition}
\label{def:semiglobal}
The function $V\in C^{1}(\mathbb{R}^{n},\mathbb{R}_{+})$ is said to be a \emph{semi-global Lyapunov function} for the system \eqref{eq:sys1} if there exists $r>0$ such that the following condition are satisfied
\begin{equation}
\label{eq:defVcondsemi}
\begin{split}
    V(0) = 0,\;\;\;\; 
    V(x)> 0,\\
    \nabla V(x)\cdot f(x) \leq 0 \text{ for }\|x\|\leq r.
\end{split}
\end{equation}
\end{definition}
Finding a semi-global Lyapunov function is usually easier than finding a global Lyapunov function. A semi-global Lyapunov function is enough to show that the equilibrium $x^{*}=0$ is robust to small perturbations which, for several engineering applications, is enough. More specifically,
\begin{definition}
We say that the (equilibrium $x^{*}=0$ of the) system \eqref{eq:sys1} is \emph{locally asymptotically stable} if it is stable and if there exists $r>0$ such that for any $\|x_{0}\|\leq r$ there exists a unique solution $x\in C^{1}([0,+\infty);\mathbb{R}^{n})$ to \eqref{eq:sys1} which satisfies in addition
\begin{equation}
\lim\limits_{t\rightarrow+\infty} x(t) = 0.
\end{equation}
\end{definition}
Similarly to global Lyapunov function, the existence of a semi-global Lyapunov function is useful to ensure local asymptotic stability
\begin{theorem}[LaSalle Invariance Principle (local)]
Assume there exists a semi-global Lyapunov function $V$, and let $S$ be the largest subset of $\{\nabla V(x)\cdot f(x) = 0\}$ invariant by the dynamics of \eqref{eq:sys1}. If $S=\{0\}$ then the system \eqref{eq:sys1} is locally asymptotically stable.
\end{theorem}
}

\section{Generation procedure}
\label{app:generation}
\subsection{Function generation}
\label{sec:fungen}
To generate random functions we sample random trees with unary and binary internal nodes, and then randomly select operators for these nodes, and variables and integers for leaves (as in \cite{lample2019,charton2020learning}). Our binary operators are the four operations and the power function. Unary operators are $\exp, \log, \mathrm{sqrt}, \sin, \cos, \tan$. 

To generate polynomials, we randomly sample a given number of monomials, with integer or real coefficients. The number of monomials, range of the coefficients, and the powers and number of terms of each monomial, are randomly selected between bounds, provided as hyperparameters.

\subsection{Backward generation}

We build globally stable systems by first generating a Lyapunov function $V$ at random, and then building a dynamic system which has $V$ as a Lyapunov function. The procedure is:

\textbf{Step 1a:} We generate $V$ as $V=V_{\text{cross}}+V_{\text{proper}}$ where $V_\text{proper}$ belongs to a given class of positive definite function and $V_{\text{cross}}$ belongs to a larger class, but of non-negative functions only with $V_{\text{cross}}(0)=0$. More specifically, we generate
\begin{equation}
 V_{\text{cross}}(x) = \sum\limits_{i=1}^{m} p_{i}^2(x),
\end{equation}
with $m$ a random integer, and $p_{i}$ random functions verifying $p_{i}(0)=0$. The nature of functions $p_{i}$ depends on the systems we want to generate (polynomial or not). Clearly $V_{\text{cross}}(x)\geq 0$ and $V_{\text{cross}}(0)=0$. We similarly generate 

\begin{equation}
V_{\text{proper}}(x) = 
\sum\limits_{i=1}^{n}\alpha_{i,j}x_{i}^{\beta_{i}}x_{j}^{\beta_{j}},
\end{equation}
with $n$ a random integer, $\beta_{i}$ random positive integers and $A = (\alpha_{i,j})_{(i,j)\in \{1,...,n\}^{2}}$ a random positive definite matrix, with a given probability of being diagonal. 
As a consequence, $V_\text{proper}$ is strictly minimal in $x=0$. When generating barrier functions, we can optionally set $V_{\text{proper}}=0$.

\textbf{Step 1b:} In this step, we increase the class of functions that can be sampled for $V_{\text{cross}}$ and $V_{\text{proper}}$ by several transformations:
\begin{enumerate}[nosep]
\item \textbf{Composition of $V_{\text{proper}}$} with probability $p_{1,c}$, replace 
\begin{equation}
    V_{\text{proper}}(x)\leftarrow I(V_{\text{proper}}(x))
\end{equation}
with $I$ selected at random from a pre-defined set of \emph{increasing-functions} (Appendix \ref{app:design_params}),
\item \textbf{Product $V_{\text{proper}}$} with probability $p_{1,m}$, replace 
\begin{equation}
V_{\text{proper}}(x) \leftarrow (V_{\text{proper}}(x)-V_{\text{proper}}(0))g(h(x)),
\end{equation}
 with $g$ selected at random from a pre-defined set of {\itshape positive-functions} (Appendix \ref{app:design_params}), $h$ a sub-expression of $V_{\text{proper}}$, i.e. $h(x)= \sum\limits_{i=1}^{q}\alpha_{\sigma(i),\sigma(j)}x_{\sigma(i)}^{\beta_{\sigma(i)}}x_{\sigma(j)}^{\beta_{\sigma(j)}}$, for $q\leq n$ and $\sigma$ a permutation of $\{1,\dots,n\}$.
\item \textbf{Composition of $V_{\text{cross}}$}: for every $i\in\{1,...,m\}$, with probability $p_{2}$, replace
\begin{equation}
p^2_{i}(x) \leftarrow {b}_{i}(\xi_{i}+ p_{i}(x)),
\end{equation}
with ${b}_{i}$ a real function that is bounded from below with a minimum (not necessarily unique) in $\xi_{i}$ and chosen at random from a pre-defined set of {\itshape bounded-functions} (Appendix \ref{app:design_params}). Recall that $p_{i}$ are the functions appearing in $V_{\text{cross}}$.
\end{enumerate}

\textbf{Step 1c:} Gathering the functions $V_{\text{proper}}$ and $V_{\text{cross}}$ together, we define the Lyapunov function (candidate) $V(x) = V_{\text{cross}}(x)+V_{\text{proper}}(x)$. Overall, we have
\begin{equation*}
V(x) = \left[I\left(\sum\limits_{i=1}^{n}\alpha_{i,j}x_{i}^{\beta_{i}}x_{j}^{\beta_{j}}\right)-I(0)\right]
g\left(\sum\limits_{i=1}^{q}\alpha_{\sigma(i),\sigma(j)}x_{\sigma(i)}^{\beta_{\sigma(i)}}x_{\sigma(j)}^{\beta_{\sigma(j)}}\right)
+\sum\limits_{i=1}^{m} b_{k}(\xi_{k}+p_{k}(x)),
\end{equation*}

where $I$ is the identity with probability $1-p_{1,c}$, $g$ is the constant function $1$ with probability $1-p_{1,m}$ and $b_{k}(x) = x^{2}$ with probability $1-p_{2}$. Such a Lyapunov function satisfies 
\begin{equation}
V(x)>V(0),\;\;\forall x\in \mathbb{R}^{n}\setminus\{0\}.
\end{equation}
Indeed, 
 \[
V(0) =\sum\limits_{i=1}^{m}b_{k}(\xi_{k})
 \]
and $V(x) > b_{k}(\xi_{k})$ for any $x\in\mathbb{R}^{n}\setminus\{0\}$, since $g$ is a positive function, $I$ is increasing and $(\alpha_{i,j})_{i,j\in\{1,...,n\}}$ is positive definite.

\textbf{Step 2:} In this step we create the random vectors orthogonal to $\nabla V$ that will be useful in the generation of the system $f$ (see Section \ref{sec:backgen}). Taking advantage of the form of the condition \eqref{eq:defVcond}, for any $x\in\mathbb{R}$, denote \[\mathcal{H}_{x}=\{z\in\mathbb{R}^{n}\;|\; z\cdot \nabla V(x)=0\}\] the hyperplane orthogonal to $\nabla V(x)$.  Then, for a random $p\in\{1,...,n\}$, generate $p$ random real-valued functions $(g_{i})_{i\in\{1,...,p\}}$, and $p$ vectors $\{e^{i}\}_{i\in\{1,...,p\}}$ from this hyperplane as follows:

\begin{equation}
    e_{j}^{i} = \begin{cases}
    A_{\tau_{2}(i)}\text{ if $j=\tau_{1}(i)$}\\
    -A_{\tau_{1}(i)}\text{ if $j=\tau_{2}(i)$}\\
    0\quad \text{otherwise},
    \end{cases}
\end{equation}
where $A=\nabla V(x)$ and $A_{j}$ refers to the $j-th$ component of the vector and $\tau_{1}$ and $\tau_{2}$ random functions from $\mathbb{N}\setminus\{0\}$ into $\{1,...,n\}$, such that $\tau_{1}(i)\neq \tau_{2}(i)$. This implies that for any $i\in\{1,...,n\}$
\begin{equation}
    \nabla V(x)\cdot e^{i} =  (\nabla V(x))_{\tau_{1}(i)}(\nabla V(x))_{\tau_{2}(i)}
    -(\nabla V(x))_{\tau_{2}(i)}(\nabla V(x))_{\tau_{1}(i)}=0.
\end{equation}

Note that, so long $\nabla V(x)\neq 0$, one can use this process to construct a generative family of $\mathcal{H}_{x}$, and the $e^{i}$ span the whole $\mathcal{H}_{x}$. If $\nabla V(x) = 0$ then $\mathcal{H}_{x}=\mathbb{R}^{n}$.

\textbf{Step 3:} Generate at random $k_{1}$ real-valued functions $(h_{i})_{i\in\{1,...,k_{1}\}}$, where $1 \leq k_{1}\leq n$ is chosen at random. Set $h_{i} = 0$ for $k_1 < i \leq n$.

\textbf{Step 4:} Build the system
\begin{equation}
f(x) = -\begin{pmatrix} h_{\pi(i)}^{2}(x)(\nabla V(x))_{i}\end{pmatrix}_{i\in\{1,...,n\}} + \sum\limits_{i=1}^{p}g_{i}(x)e^{i}(x),
\end{equation}
with $\pi$ a random permutation of $\{1,...,n\}$. 
\\

Overall, the function $f$ satisfies
\begin{equation}
    \nabla V (x) \cdot f(x) = - \left(\sum\limits_{i=1}^{n} h_{\pi(i)}^{2}(x)(\nabla V(x))_{i}^{2}\right)\leq 0,
\end{equation}
hence $V$ is a Lyapunov function of the system 
$\dot x(t) = f(x(t))$.

\textbf{Step 5:} Expand and simplify the equations of $f$ (using Sympy), in order to eliminate obvious patterns due to the generation steps (that the model could recognize and leverage), eliminate duplicated systems in the training set, and limit the length of training sequences. All polynomial systems are expanded into normal form.

\subsection{Backward generation modes}

\textbf{Polynomial generation}: we generate polynomial systems with sum-of-square Lyapunov functions to allow for easy comparison with existing methods such as SOSTOOLS \cite{prajna2002introducing,prajna2005sostools}. In this case, all $P_{i}$ are polynomials with no zero-order term and $p_{1,c}=p_{1,m}=p_{2}=0$. Also, $f_{i}$ and $g_{i}$ are polynomials (Appendix~\ref{sec:fungen}). We generate $f_{i}$ with a degree lower or equal to half the maximal degree of $g_{i}$ and a maximal value of coefficients of the order of the square root of the maximal value of $g_{i}$. Since the $f_{i}$ are squared in the final system, this allows $f_{i}^{2}$ and $g_{i}$ to have the same order, and prevents the transformer from inferring unwanted additional information by looking at the higher degree monomial.

\textbf{Generic generation}: $P_{i}$ is generated as
    $P_{i}(x) = Q_{i}(x)-Q_{i}(0)$,
    where $Q_{i}(x)$ is a random function generated as per Appendix~\ref{sec:fungen} 
    and $f_{i}$ and $g_{i}$ are also generated as per Appendix~\ref{sec:fungen}. Optionally the functions can be generated as polynomials of non-polynomial operators taken from a pre-defined set of \emph{operators}.

\textbf{Other generation modes}: we have other generation modes corresponding to interesting particular cases: gradient flow systems, systems where the 2-norm (resp. a weighted 2-norm) is a Lyapunov function, etc.

\subsection{Generation design parameters}\label{app:design_params}

Our generator allows us to generate generic stable systems and yet to have a large control on the distribution. For polynomials, for instance, we have a control on the maximal and average degree, number of monomials, power and number of variables of the monomials, coefficients, etc. We can also specify whether the coefficients are integers, floats, with which precision. Overall we have a total of 36 generation hyper-parameters that influence the distribution of the synthetic data created. The main generation design parameters are:
\begin{itemize}[nosep]
        \item int\_base: encoding base for integers
        \item max\_int: Maximum integer value
        \item precision: Float numbers precision
        \item prob\_int: Probability of sampling integers vs variables (for non-polynomial expressions)
        \item min\_dim: minimal number of equations in the system
        \item max\_dim: maximal number of equations
        \item max\_degree: maximal degree of polynomial terms in a Lyapunov function
        \item n\_terms: maximal number of terms in polynomials for the Lyapunov function
        \item nb\_ops\_proper: maximal number of operators in $V_{proper}$ (non polynomial generation)
        \item nb\_ops\_lyap: maximal number of operators in $V_{proper}$ (non polynomial generation)
        \item operators\_lyap: list of operators to be considered (non polynomial generation)
        \item polynomial\_V: if true generated expressions are polynomials of (potentially non-polynomial) operators
        \item pure\_polynomial: generate polynomial systems only
        \item cross\_term: $V_{cross}= 0$ if False.
        \item max\_nb\_cross\_term: bound on m in $V_{cross}$
        \item proba\_diagonal: with this probability, the positive definite form of $V_{proper}$ is imposed to be diagonal
        \item only\_2\_norm: if True, the Lyapunov function is the 2-norm.
        \item strict: if True, generates a strict Lyapunov function (i.e. $\nabla V\cdot f<0$)
        \item proper: if set to false, $V_{proper}=0$ and $V$ is only a barrier function.
        \item float\_resolution\_poly: float resolution of the polynomials generated by generate\_bounded\_polynomial.
        \item generate\_gradient\_flow: When set to True, the backward generation only generates gradient flows systems.
        \item gen\_weight: exponential weight which bias the sampling of $k_{1}$ and $p$, the number of components of non-zero $h_{i}$ and $g_{i}$.
        \item max\_order\_pure\_poly: maximal polynomial order of $h_{i}$
        \item max\_n\_term\_fwd: maximal number of terms in each equations in the fwd generation
        \item SOS\_checker: if True, uses a SOS verifier to evaluate the candidate Lyapunov function (if False uses the verifier based on \texttt{shgo})
        \item SMT\_checker: if True, uses an SMT verifier to evaluate the candidate Lyapunov function (if False uses the verifier based on \texttt{shgo})
        \item multigen: number of different system generated per Lyapunov function.
        \item increasing\_func: the set of increasing functions used in the generation (see Step 1b). Default is $\{\exp, \ln(1+x^{2}), \sqrt{1+x}\}$.
        \item positive\_func: the set of positive functions used in the generation (see Step 1b). Default is $\{\exp, 1+\cos(x), 1+\sin(x)\}$.
        \item bounded\_func: the set of bounded functions used in the generation (see Step 1b). Default is $\{\cos, \sin\}$.
\end{itemize}

\subsection{Forward SOS solver}
\label{sec:SOS}
SOSTOOLS is one of the most famous toolbox for sum-of-square optimization, in particular for finding SOS Lyapunov functions \cite{prajna2002introducing,prajna2005sostools}. It is natively available in MATLAB and relies on an underlying SDP solver that can be chosen. In Python an analogous toolbox is the package SumOfSquares \cite{Yuan2024} which relies on the same principle, however does not have specific functionalities for Lyapunov functions. As a consequence we implemented these functionalities in our codebase based on the MATLAB implementations in SOSTOOLS. We implemented a function \texttt{SOS\_checker}, which takes in input a system of equations in sympy and a candidate Lyapunov function and checks SOS conditions on $V(x)$ and $-\nabla V(x)\cdot f(x)$, and a function \texttt{findlyap}, analogous to the \texttt{findlyap} function in SOSTOOLS, which takes a system of equations in sympy and either returns a function satisfying  SOS conditions on $V(x)$ and $-\nabla V(x)\cdot f(x)$, returns false if no such function exists, or returns none if it fails to provide an answer. SumOfSquares relies itself on picos \cite{PICOS} and we use the default solver cvxopt \cite{AndersenDahlVandenberghe2023}.

\subsection{List of datasets}\label{app:datasets}

\begin{table}[h]
    \small
    \centering
    \begin{tabular}{l|ccc}
        \toprule 
        Dataset & Description & Size & Resources \\
        & & (000) & (CPU.hours) \\
        \midrule
        BPoly & Backward polynomial systems, non-zero& 1,000 & 210 \\
        BNonPoly & Backward non-polynomial systems, non-zero& 1,000 & 220 \\
        \midrule 
        FBarr &Forward, non-homogeneous polynomial barrier functions & 300 & 9,670 \\
        FLyap & Forward, homogeneous polynomial Lyapunov functions & 100 & 4,620 \\
        FSOSTOOLS& Forward, SOSTOOLS solved systems& 1.5 &  \\
         \bottomrule
    \end{tabular}
    \caption{\small \textbf{Datasets generated.} Backward systems are degree 2 to 5, forward systems degree 2 to 3. All forward systems are polynomial.}
    
   \label{tab:datasets}
\end{table}

\section{Additional results}\label{app:suppl_results}

\subsection{Impact of multigeneration}
\label{sec:multigen}

In the backward generation procedure, after sampling one random $V$, it is possible to generate any number of different systems $f_i$ such that $V$ is the Lyapunov function for each of the systems $f_i$. We call the maximal number of system generated per Lyapunov function the \texttt{multigen} parameter. The actual number of systems generated per Lyapunov function is chosen at random for each Lyapunov function between 1 and \texttt{multigen}. In Section \ref{main_results} we reported results using \texttt{multigen} equal to 5. Here we report the in-domain and out-of-domain performance of the models trained on backward BPoly datasets of size 1 million varying the parameter \texttt{multigen}.

\begin{table}[h]

\centering
\small
\begin{tabular}{ccc}
\toprule
& In-domain & OOD \\
Multigen & BPoly & FLyap\\
\midrule
1 & 95 & 58 \\
5 & 100 & 73 \\
10 & 100 & 75 \\
25 & 100 & 76 \\
50 & 100 & 70 \\
100 & 100 & 68 \\
\bottomrule
\end{tabular}
\caption{\small \textbf{In-domain and out-of-domain accuracy of models}. Beam size 50. }
\label{multigen_impact}
\end{table}

Table \ref{multigen_impact} shows that generating a moderate amount of different systems with the same Lyapunov function actually improves the model capability to generalize out-of-domain. This suggests that the model is learning, at least partially, to separate the parts of the system which contribute to the Lyapunov function. Above a certain multigen threshold, model performances start to decline. This may be due to the low diversity present in the dataset, i.e. the limited number of different Lyapunov functions the model is trained on (the total number of systems in the training set remains constant so the total number of Lyapunov function decreases with the value of the parameter \texttt{multigen}).

\subsection{Performance of smaller transformer models}

In Section \ref{main_results} we report results using a transformer with 8 encoder and decoder layers, 10 attention heads and an embedding dimension of 640. 
We also trained smaller models with 6 encoder and decoder layers, 8 attention heads and an embedding dimension of 512. Tables \ref{ablative_1}, \ref{ablative_2} report the main results. Results are in line with what we showed in section \ref{main_results}

\begin{table}[h]

\centering
\small
\begin{tabular}{lcc||lcc}
\toprule
& In-domain & OOD & & In-domain & OOD\\
Backward datasets & & FLyap & Forward datasets & & BPoly \\
\midrule
BPoly (polynomial) & 100 & 70 & FBarr (barrier) & 97 & 13 \\
BNonPoly (non-poly) & 85 & 71 & FLyap (Lyapunov) & 86 & 11 \\
\bottomrule
\end{tabular}
\caption{\small \textbf{In-domain and out-of-domain accuracy of models}. Beam size 50. }
\label{ablative_1}
\end{table}

\begin{table}[h]

\centering
\small
\begin{tabular}{lr|ccccc}
\toprule
Forward & Mixing & & & \multicolumn{2}{c}{Into the wild}  \\
datasets & proportion & FBarr & FLyap & Poly3 & Poly5 \\
\midrule
No mixture & 0\% & 31 & 70 & 2.3 & 1.6 \\
\midrule
FBarr & 0.01\% & 60 & 72 & 2.5 & 1.7\\
 & 0.1\% & 93 & 72 & 9.2 & 6.4 \\
\midrule
FLyap & 0.01\% & 29 & 73 & 2.8 & 1.6\\
 & 0.1\% & 19 & 76 & 2.9& 1.7\\
\bottomrule
\end{tabular}
\caption{\small \textbf{Performance of mixing backward data (BPoly) with a small number of forward examples on forward benchmark and ``into the wild''}. Beam size 50.}
\label{ablative_2}
\end{table}

\section{Comparison of SOS, SMT and  \texttt{shgo}}\label{app:checker}

We compare our model performance when we employ them to discover new Lyapunov function. We report performances with dReal SMT and SOS verifiers for Poly and dReal SMT and \texttt{shgo} for NonPoly distributions, respectively. Table \ref{intothewild_performance_two_checkers} shows that SMT results are slightly lower, because of timeouts (which we report in Table \ref{timeout_sos_intothewild}, but comparable. Note that the performances on polynomial systems were already theoretically guaranteed thanks to the former SOS verifier.

\begin{table}[h]
    \small
    \centering
    \begin{tabular}{lccc}
        \toprule
         Model (by training distribution) & FBarr & BPolyMixture & NonPolyMixture \\
        
        \midrule
        Poly3 & 10.5/11.7 & 11.1/11.8 & 10.6/11.2\\
        Poly5 & 6.5/8.0 & 8.7/10.1 & 8.4/9.9\\
        NonPoly & & & 8.3/12.7\\
        \bottomrule
    \end{tabular}
    \caption{\small \textbf{Results of SMT} with SOS and \texttt{shgo} verifiers for Poly and NonPoly systems, respectively. 
    }
    \label{intothewild_performance_two_checkers}
\end{table}

\begin{table}[!h]
    \small
    \centering
    \begin{tabular}{lcc}
        \toprule
        Test sets & \multicolumn{2}{c}{Into the wild} \\
        Timeout &10 minutes &120 minutes \\
        \midrule
        Correct Lyap function & 87.3 & 92.2\\
        SMT Timeouts & 10.8 & 5.8 \\
        Incorrect Lyap function & 1.9 & 2.0  \\
        \bottomrule
    \end{tabular}
    \caption{\small \textbf{SMT timeout and error rates.} Most SMT failures are due to timeout.}
    \label{timeout_sos_intothewild}
\end{table}

\section{AI method sweep}\label{app:sweep}

To report the AI-based tools results on the seven benchmarks (BPoly, BNonPoly, FLyap, FBarr, Poly3, Poly5, NonPoly) we did a hyperparameter sweep. To get the best hyperparameter setting, we sweep on FLyap and then fix these hyperparameters for the different datasets. In bold we show the chosen parameters, selected to maximize the correctness on FLyap, subject to the 20 minutes timeout.

\textbf{Lyznet \citep{lyznet}}
\begin{itemize}[nosep]
    \item lr = [$3\cdot10^{-5}$, $\mathbf{10^{-4}}$, $3\cdot10^{-4}$]
    \item points = [\textbf{100,000}, 300,000, 1,000,000]
    \item layer width = [(2,20), \textbf{(3,6)}, (6,2)]
    \item epoch = [1, 5, \textbf{25}]
    \item net type = [None, \textbf{Poly}]
\end{itemize}
\newpage
\textbf{Fossil 2.0 \citep{fossil2}}
\begin{itemize}[nosep]
    \item iters = [10, \textbf{50}, 250]
    \item activations = [($\mathbf{x^2}$), ($x^2$,$x^2$), (sigmoid), (sigmoid, sigmoid), ($\textrm{poly}_4$), ($\textrm{poly}_4$, $\textrm{poly}_4$)]
    \item hidden neurons = [6, \textbf{10}, 20]
    \item data = [500, \textbf{1000}, 2000]
    \item lr = [0.01, 0.03, \textbf{0.1}]
\end{itemize}

\textbf{ANLC v2 \citep{grande2023augmented}}
\begin{itemize}[nosep]
    \item iters = [10, \textbf{50}, 250]
    \item activations = [($\mathbf{x^2}$, $\mathbf{x}$, $\mathbf{x}$), ($x^2$, $x^2$, $x$), ($x^2$, $x^2$, $x$, $x$), ($x^2$, $x^2$, $x^2$, $x$)]
    \item hidden neurons = [6, \textbf{10}, 20]
    \item max data = [500, \textbf{1000}, 2000]
    \item lr = [\textbf{0.01}, 0.03, 0.1]
\end{itemize}

\section{Some examples}
\label{app:examples}
To understand the model performance and compare against the SOSTOOL performance, we manually inspect some systems with 2 or 3 equations where the following conditions hold: (1) the Jacobian of the system has the maximum eigenvalue with real part equal to 0 (i.e. tools like the spectral mapping theorem cannot decide on the stability), (2) no weighted 2-norm functions can be a Lyapunov function, (3) \texttt{findlyap} times out after 4 hours. We show some examples below. 

\subsection{A polynomial system with non polynomial solution}

\textbf{System}

\begin{equation*}
\begin{cases}
\dot x_0 &= -x_0+x_0x_1 \\
\dot x_1 &= -x_1
\end{cases}
\end{equation*}

It's known that there is no polynomial Lyapunov function for this system ~\citep{ahmadi2011}. Our poly models and \texttt{findlyap} failed, as expected. Nonetheless, one of our non-poly models with beam search of beam size $100$ proposed $V(x)=\ln(1+5x_0^2) + x_1^2$ similar to the one that was recently found in~\citep{ahmadi2011}.

It's clear that $V(0)=0$ and $ V(x)> 0$ for all $x \neq 0$. Also 
\begin{equation}
    \begin{split}
 V(x) \cdot f(x) &= 
 \frac{-10x_0^2+10x_0^2x_1-2x_1^2(1+5x_{0}^{2})}{1+5x_0^2}\\
 &=\frac{-5x_0^2 - 5x_0^2x_1^2-5(x_0-x_0x_1)^2-2x_1^2}{1+5x_0^2}\leq 0       
    \end{split}
\end{equation}
as desired.

\subsection{A system that has no diagonal Lyapunov function}

\textbf{System}

\begin{equation*}
\begin{cases}
\dot x_0 &= 2x_1^2 \\
\dot x_1 &= -10x_1
\end{cases}
\end{equation*}

\textbf{Model inference}: Our model recovers $V(x) = 10x_0^2+2x_0x_1^2+3x_1^4+6x_1^2$. 

Clearly $V(0) = 0$ and $V(x) = 9(x_0)^2 + (x_0+x_1^2)^2 + 2(x_1^2)^2 + 6x_1^2 > 0$ for all $x \neq 0$. 
Also $ \nabla V(x) \cdot f(x) = -x_1^2(116x_1^2+120) \leq 0$.

\textbf{Non existence of a Diagonal Lyapunov function}: Suppose for the sake of contradiction that there exists a function $V_1$ which satisfies \ref{eq:defVcond} and can be expressed as \[ V_1(x) = \sum_{i=1}^{n} a_i x_0^i + \sum_{j=1}^{m} b_j x_1^j.\]

Clearly $V_1(0) = 0$. Given that $V_1(x_0, 0) > 0$ for $x_0 \neq 0$, it follows that $n$ is even and $a_n > 0$. 
Also we know that $\displaystyle \nabla V_1(x) \cdot f(x) = 2\sum_{i=1}^{n} ia_i x_0^{i-1}x_1^2 - 10\sum_{j=1}^{m} jb_j x_1^j \leq 0$ for all choices of $(x_0, x_1)$. 
If we let $x_1 = 1$ we obtain $\displaystyle \nabla V_1(x) \cdot f(x) = 2\sum_{i=1}^{n} ia_i x_0^{i-1} - 10\sum_{j=1}^{m} jb_j$. This expression can be seen as a polynomial $g(x_0)$ with real coefficients and odd degree $n-1$. The leading coefficient, $2na_n$, is positive because $a_n > 0$ and $n \geq 1$. This means that $\displaystyle \lim_{x_0 \to +\infty} g(x_0) = +\infty$, meaning that there exists an $x_0$ such that $g(x_0) > 0$. This contradicts \ref{eq:defVcond}. 

\subsection{A system with 3 equations and a higher degree}

\textbf{System}

\begin{equation*}
\begin{cases}
\dot x_0 &= -7x_0^5-4x_0^3x_1^2-5x_0^3 \\
\dot x_1 &= 7x_0^4-3x_1-2x_2 \\
\dot x_2 &= -8x_0^2-9x_2
\end{cases}
\end{equation*}

\textbf{Model inference}: Our model recovers different solutions. Here we show two of them

\begin{gather*}
V_1(x) = 4x_0^4+10x_0^2x_1^2+2x_0^2x_1+10x_0^2x_2^2-4x_0^2x_2+20x_0^2+10x_1^2x_2^2+4x_1^2-2x_1x_2+8x_2^4+4x_2^2,\\
V_2(x) = 2x_0^4 + 2x_0^2x_1^2 + 3x_0^2 + 2x_1^2 + x_2^2.
\end{gather*}

We checked with SumOfSquares that $V_1 > 0, V_2 > 0, \nabla V_1 \cdot f \leq 0$ and $\nabla V_2 \cdot f \leq 0$.

\subsection{Other examples}

\begin{table}[h]
\small
\centering
\begin{tabular}{l|l}
\toprule
System & Lyapunov function \\
\midrule
$\begin{cases} 
\dot{x_0} = -5x_0^3-2x_0x_1^2\\ 
\dot{x_1} = -9x_0^4+3x_0^3x_1-4x_1^3
\end{cases}$
& $V(x) = 6x_0^6+7x_0^4+x_0^3+10x_0^2+8x_1^2 $ \\
\midrule
$\begin{cases} 
\dot{x_0} = -x_0^5-4x_0^3-9x_0x_1^4+3x_0x_1^3\\ 
\dot{x_1} = -3x_0^4x_1^2-10x_0^3x_1+3x_0x_1^2-7x_1^3
\end{cases}$
& $V(x) = x_0^4+9x_0^2+3x_1^2$ \\
\midrule
$\begin{cases} 
\dot{x_0} = -3x_0^3+3x_0x_2-9x_0\\ 
\dot{x_1} = -x_0^3-5x_1+5x_2^2\\
\dot{x_2} = -9x_2^3
\end{cases}$
& $V(x) = x_0^4+7x_0^2x_2^2+3x_0^2+4x_0x_2^2+3x_1^2+2x_2^4+10x_2^2$ \\
\midrule
$\begin{cases} 
\dot{x_0} = -8x_0x_1^2-10x_1^4\\ 
\dot{x_1} = -8x_1^3+3x_1^2-8x_1\\
\dot{x_2} = -x_2
\end{cases}$
& $V(x) = 4x_0^2-2x_0x_1^2+6x_1^4+4x_1^2+x_2^2$ \\
\bottomrule
\end{tabular}
\caption{\small \textbf{Some additional examples generated from our models}.}
\label{additional_examples}
\end{table}

\end{document}